\pdfoutput=1

\documentclass[11pt]{article}

\usepackage[]{acl}

\usepackage{times}
\usepackage{latexsym}

\usepackage[T1]{fontenc}

\usepackage[utf8]{inputenc}

\usepackage{microtype}

\usepackage[whole]{bxcjkjatype}
\usepackage{booktabs}
\usepackage{graphicx}
\usepackage{multirow}
\usepackage{enumitem}
    \setlist{itemsep=0pt, topsep=3pt}
\usepackage{pifont}
\usepackage{array}
\usepackage{subcaption}
\usepackage{float}
\usepackage{multicol}

\newcommand{\entq}{\textsc{EntQ}}
\newcommand{\cntq}{\textsc{CntQ}}
\newcommand{\certainty}{\textit{Certainty}}
\newcommand{\variety}{\textit{Variety}}

\title{N-best Response-based Analysis of Contradiction-awareness in \\Neural Response Generation Models}

\author{
Shiki\,Sato$^{1}$\hspace{1em}
Reina\,Akama$^{1,3}$\hspace{1em}
Hiroki\,Ouchi$^{2,3}$\hspace{1em}
Ryoko\,Tokuhisa$^{1}$\hspace{0em} \\
\textbf{Jun\,Suzuki}$^{1,3}$\hspace{1em}
\textbf{Kentaro\,Inui}$^{1,3}$\\[3pt]
$^{1}$Tohoku University\hspace{1em}
$^{2}$Nara Institute of Science and Technology\hspace{1em}
$^{3}$RIKEN\hspace{1em}
\\\texttt{\{shiki.sato.d1,akama,tokuhisa,jun.suzuki,inui\}@tohoku.ac.jp}
\\\texttt{hiroki.ouchi@is.naist.jp}
}

\begin{document}
\maketitle
\begin{abstract}
Avoiding the generation of responses that contradict the preceding context is a significant challenge in dialogue response generation. 
One feasible method is post-processing, such as filtering out contradicting responses from a resulting $n$-best response list.
In this scenario, the quality of the $n$-best list considerably affects the occurrence of contradictions because the final response is chosen from this $n$-best list.
This study quantitatively analyzes the contextual contradiction-awareness of neural response generation models using the consistency of the $n$-best lists.
Particularly, we used polar questions as stimulus inputs for concise and quantitative analyses.
Our tests illustrate the contradiction-awareness of recent neural response generation models and methodologies, followed by a discussion of their properties and limitations.
\end{abstract}


\section{Introduction}  \label{sec:introduction}
Recent advanced response generation models~\cite{zhang:acl2020demo:dialogpt, adiwardana:arxiv2020:meena, roller:eacl2021:blenderbot} can generate relevant and meaningful responses, which can resolve dull response problems~\cite{vinyals:icml2015ws:neuralconv, sordoni:naacl2015:gen-context-sensitive, serban:aaai2016:HRED}.
This advancement reveals additional flaws in the quality of neural model responses, such as \emph{contradiction}.
Contradiction is a critical error in dialogue because a single contradictory response can disrupt the flow of the dialogue~\cite{higashinaka:sigdial2015:taxonomy}.

A generation model outputs a response by selecting the candidate with the highest likelihood ($1$-best) from an $n$-best candidate list.
Prior work has demonstrated that generating the $n$-best lists with noncontradictory $1$-bests is an open challenge~\cite{nie:acl2020:i-like-fish, kim:emnlp2020:will-i-sound-like-me, li:acl2021:addressing}.
Thus, one practical technique for avoiding contradiction is to have an accurate contradiction detector that eliminates all contradictory candidates from the $n$-best list~\cite{nie:acl2020:i-like-fish}.
In this scenario, the consistency of all candidates in the $n$-best list, not just the $1$-best, substantially impacts whether the final output is contradictory because the final response is chosen from the $n$-best list.
Nonetheless, earlier quantitative investigations of contradiction relied solely on $1$-bests from models~\cite{li:acl2021:addressing}.

\begin{figure}[t]
    \begin{center}
        \includegraphics[width=\columnwidth]{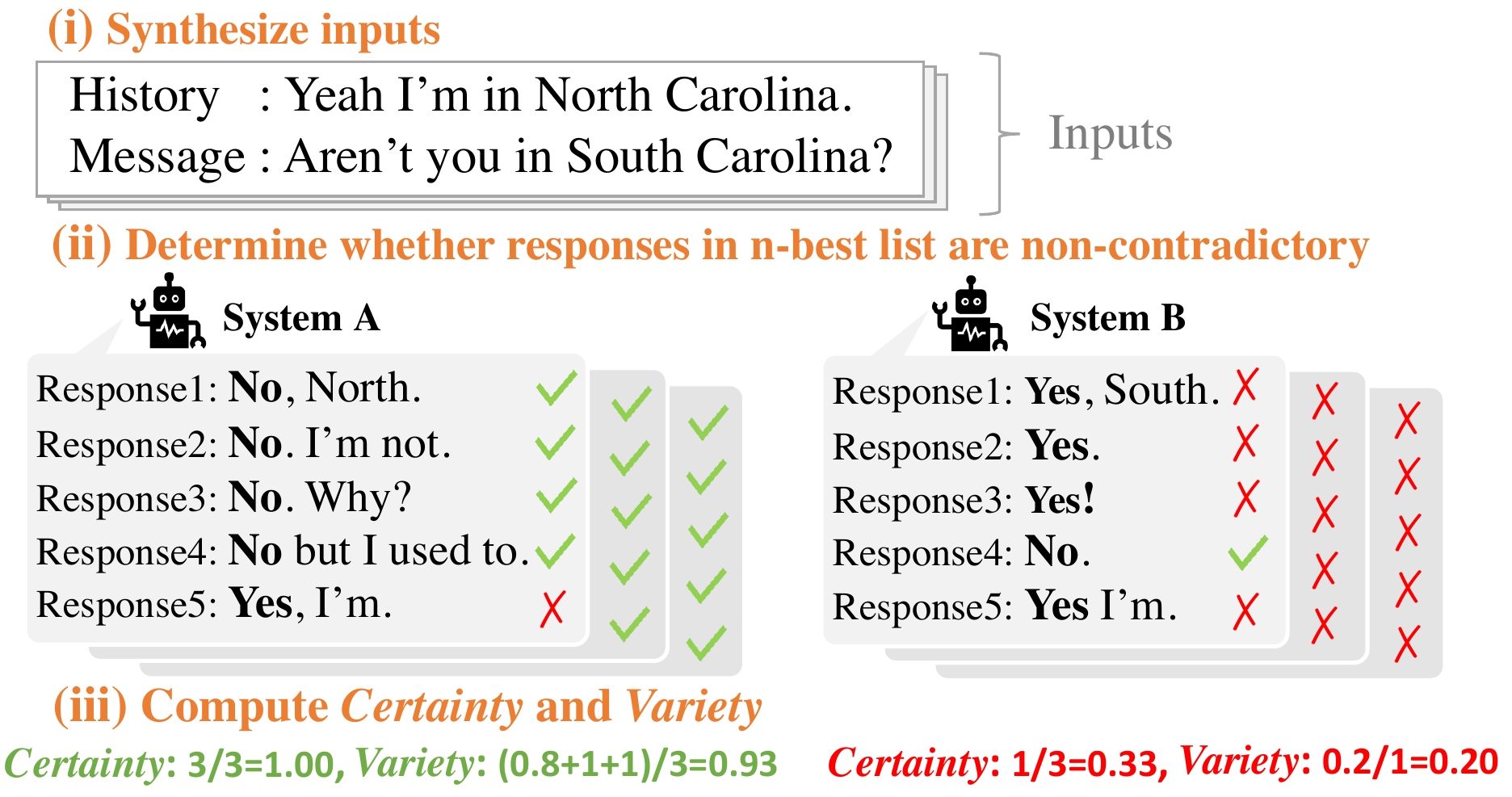}
    \end{center}
    \vspace{-1mm}
    \caption{Overview of our analysis framework. The framework analyzes $n$-best lists by (i) synthesizing a stimulus input that induces contradictions, (ii) automatically determining whether responses in the $n$-best lists are contradictory, and (iii) computing \certainty{} and \variety{}.}
    \label{fig:overview}
    \vspace{-3mm}
\end{figure}

In this study, we analyze the $n$-best lists generated by the models to explore methods for enhancing neural response generation to avoid contradiction.
Specifically, we first consider how analyzing an $n$-best list should be approached. 
Then, we propose a method for statistically analyzing the $n$-best lists (Figure~\ref{fig:overview}). 
Since it is impractical to study all conceivable contradictions in a dialogue, we first focus on contradictions in response to polar questions.%
    \footnote{Codes and test set are available at\\ \url{https://github.com/shiki-sato/nbest-contradiction-analysis}}
We use our method to highlight the contradiction-awareness of recent high-performance neural response generation models and methodologies.
Our results show that beam search has limitations in terms of avoiding contradiction and that the newer techniques, such as unlikelihood training~\cite{welleck:iclr2020:unlikelihood}, can help overcome these limitations.

\begin{table*}[t]
    \centering
    \small
    \tabcolsep 5pt
    \begin{tabular}{lrl l lrl}
        \toprule
            \multicolumn{3}{l}{NLI data} 
            && \multicolumn{3}{l}{Dialogue context for our test} 
        \\
        \cmidrule{1-3} \cmidrule{5-7}
            \textbf{Entailment}
            & Premise: & yeah i'm in North Carolina 
            & \multirow{2}{*}{$\longrightarrow$}
            & \textbf{\entq{}}
            & History: & Yeah I'm in North Carolina. 
        \\
            & Hypothesis: & I'm in North Carolina. 
            &&
            & Message: & Are you in North Carolina?
        \\
        \cmidrule{1-3} \cmidrule{5-7}
            \textbf{Contradiction}
            & Premise: & yeah i'm in North Carolina
            & \multirow{2}{*}{$\longrightarrow$}
            & \textbf{\cntq{}}
            & History: &Yeah I'm in North Carolina.
        \\
            & Hypothesis: & I'm in South Carolina.
            &&
            & Message: & Aren't you in South Carolina? 
        \\
        \bottomrule
    \end{tabular}
    \caption{Acquiring dialogue context by transforming the Natural Language Inference (NLI) data.}
    \label{tab:nli-transform}
\end{table*}

\section{Analysis perspectives}
\label{sec:analysis-point}
First, $n$-best lists must be generated to prevent contradiction, assuming the filters can remove contradictory responses.
An ideal model produces output that is noncontradictory and outperforms in many other criteria, such as relevance or informativeness.
A model must generate at least one noncontradictory candidate to deliver a noncontradictory output.
Furthermore, even noncontradictory candidates could be eliminated based on other criteria (e.g., relevance, informativeness).
Therefore, it can be hypothesized that having more noncontradictory responses in an $n$-best list would enhance the final output quality across various criteria.
Taking the above into account, we examine $n$-best lists based on the certainty of the existence of noncontradictory responses (\textbf{\certainty{}}), and the variety of noncontradictory responses (\textbf{\variety{}}):
\begin{itemize}
    \setlength{\leftskip}{-5pt}
    \item \textbf{\certainty{}}: The proportion of the $n$-best lists that have at least one noncontradictory response.
    \item \textbf{\variety{}}: The proportion of noncontradictory responses in each $n$-best list when only the $n$-best lists with at least one noncontradictory response are collected.
\end{itemize}
Given a set of inputs $\mathcal{Q}$, we calculate them as follows:

\begin{displaymath}
    \textbf{\certainty{}}=\frac{|\mathcal{Q}'|}{|\mathcal{Q}|},
    \textbf{\variety{}}=\frac{1}{|\mathcal{Q}'|}\sum_{q \in \mathcal{Q}'}\frac{\mathrm{cnt}(f(q))}{|f(q)|}
\end{displaymath}
\begin{displaymath}
    \mathcal{Q}' = \{q \mid \mathrm{cnt}(f(q)) > 0 , q \in \mathcal{Q}\}
\end{displaymath}
where $f(\cdot)$ is an $n$-best list generation function and $\mathrm{cnt}(\cdot)$ is a function that returns the number of noncontradictory responses from a given $n$-best list.
For example, the \certainty{} of a model that generates $n$-best lists with a combination of noncontradictory and contradictory responses is high, but its \variety{} is low. 
However, a model that always generates $n$-best lists with only noncontradictory or contradictory responses has a high \variety{} but a low \certainty{}.
We anticipate that $n$-best lists must include noncontradictory responses (\certainty$=1.0$), with a high proportion (high \variety{}).


\section{Analytical inputs and evaluation}
\label{sec:analysis-method}
To analyze a model from the aforementioned viewpoints, we consider how to prepare the analytical inputs and evaluate the generated responses in this section.

\subsection{Inputs for highlighting contradictions}
\label{subsec:input-preparation}

\paragraph{Polar echo question.}
An \textit{echo question}~\cite{noh:lp1998:echo} confirms or clarifies the context information by repeating the utterance of another speaker.
It is commonly used when the speaker did not hear or understand what was said correctly, or when the speaker wishes to express incredulity.
Based on \citet{li:acl2021:addressing}'s discovery, contradictions emerge mostly when speakers refer to earlier information communicated in dialogue; we use echo questions as stimulus input in our analysis to elicit contradictory responses.
We use polar-typed echo questions to make our analysis more succinct and quantitative. 
Since polar questions allow for basically only two responses, \textit{yes} or \textit{no}, we can clearly determine whether the generated response is contradictory or not.
Furthermore, by analyzing the produced responses as a yes/no binary classification issue, it allows for quantitative discussion of experimental outcomes based on the probability level.

\paragraph{Input preparation.}
We use the dataset from the natural language inference (NLI) task to effectively obtain the analytical inputs described in the preceding paragraph.
This dataset specifies the logical relationship (i.e., entailment, neutrality, or contradiction) between a premise and its associated hypothesis.
We transform the NLI dataset into dialogue data using a set of basic rewriting rules.%
    \footnote{The details are described in Appendix~\ref{sec:writing-rule}.}
Our test involves two types of inputs, which can be classified as follows:
\begin{itemize}
    \setlength{\itemsep}{-3pt}
    \setlength{\leftskip}{-5pt}
    \item \entq{}: generating a \emph{confirmation} response.
    \item \cntq{}: generating a \emph{refutation} response.
\end{itemize}
Table~\ref{tab:nli-transform} displays the input samples and how they are transformed from the initial NLI data.
Each input is made up of the following two utterances: the history and message.
In our analysis, the model generates responses to a given message, assuming the model has generated the history in the preceding turn.

\subsection{Contradiction detection for output}
\label{subsec:detection-contradiction}
To compute the \certainty{} and \variety{}, we must first determine whether each generated response in the $n$-bests compared to the inputs is contradictory.
The simplest method for detecting the contradictions is to check whether the response begins with \textit{yes} or \textit{no}.
However, in the event of an indirect expression (e.g., \textit{Why not?}), this method cannot detect the contradictions.
Therefore, we use an automated yes-no classifier to categorize the $n$-best responses to \entq{}/\cntq{}.
We train the classifier by fine-tuning RoBERTa~\cite{liu:arxiv2019:roberta} using the Circa dataset~\cite{louis:emnlp2020:circa}, which comprises pairs of polar questions and indirect responses, as well as annotations for the answer's interpretation, to categorize utterances as affirmations or refutations.%
\footnote{The details are described in Appendix~\ref{sec:classifier}.}


\section{Experiments}
\label{sec:experiments}
We demonstrate how our framework shows the properties of $n$-best lists, which could be quite influential in terms of avoiding contradiction.
We demonstrate this by comparing the $n$-bests generated by conventional beam search (BS) versus recently proposed techniques.

\subsection{Experimental settings} \label{subsec:experimental-settings}
\paragraph{Inputs preparation.}
We used the Multi-Genre NLI Corpus
~\cite{williams:naacl2018:challengecorpus} to obtain analytical inputs, which is a large scale and is consistent in good quality NLI data.
We created $2,\!000$ \entq{}/\cntq{} inputs by extracting $2,\!000$ samples labeled with \textit{entailment} or \textit{contradiction}.%
\footnote{We used the samples in the \textsc{Telephone} domain; this domain covers open-domain conversations.}

\paragraph{Response generation models.}
We used the following two recently developed high-performance models: DialoGPT~\cite{zhang:acl2020demo:dialogpt} and Blender~\cite{roller:eacl2021:blenderbot}.%
\footnote{The details of the settings are described in Appendix~\ref{sec:details-analysis}.}

\subsection{Analysis of \textit{n}-best using beam search}
\label{subsec:analysis-beamsearch}
\begin{figure}[t]
    \begin{center}
        \includegraphics[width=\columnwidth]{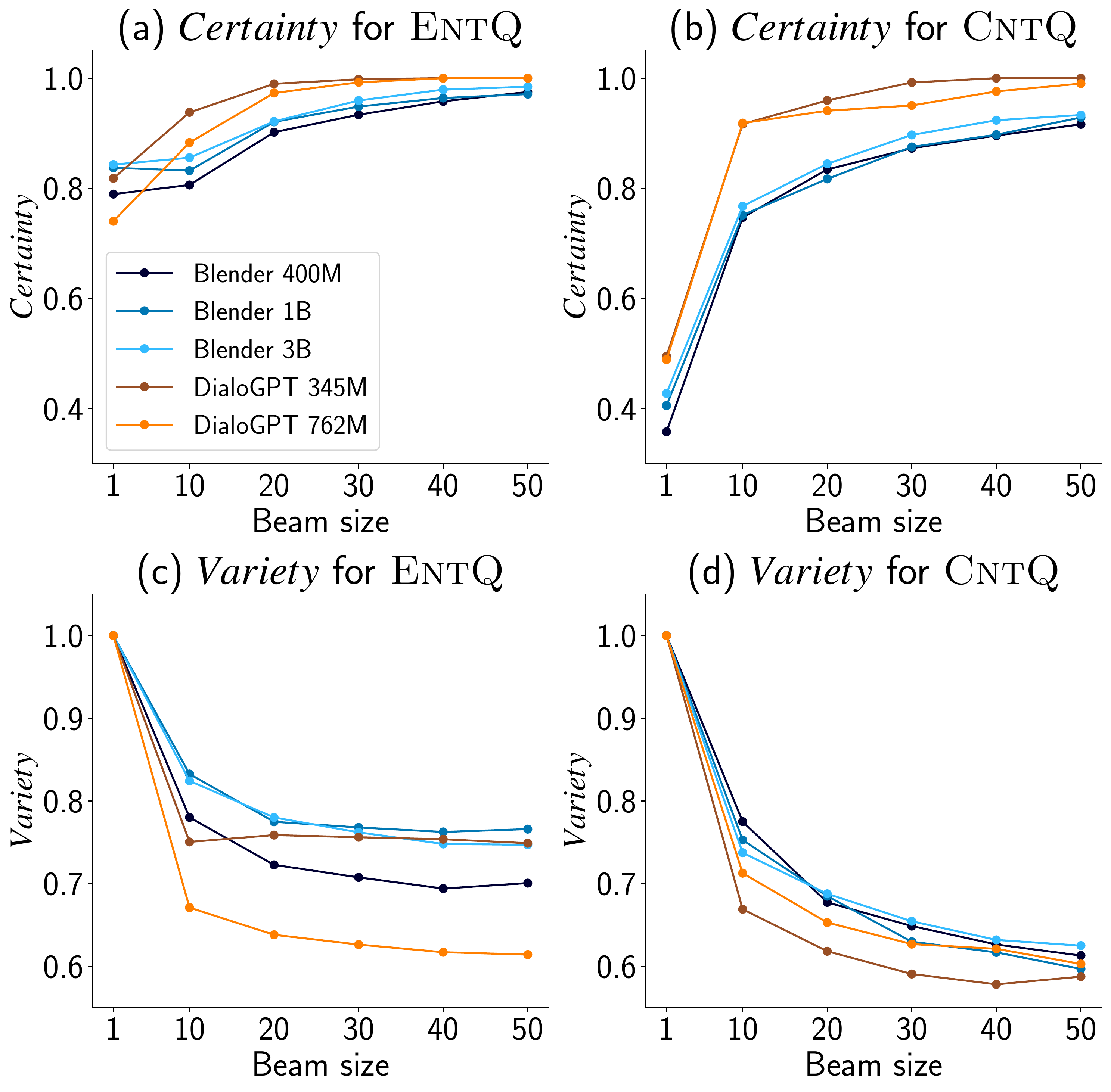}
    \end{center}
    \caption{\certainty{} and \variety{} of $n$-best lists using beam search with various beam sizes.}
    \label{fig:usual-beam}
\end{figure}

Let $B$ denote the beam size during generation.
It has been empirically found that using beam search with $B=10$ to generate a response yields excellent quality results and has a frequently used value~\cite{zhang:acl2020demo:dialogpt, roller:eacl2021:blenderbot}.
Table~\ref{tab:usual-beam10} displays the \certainty{} and \variety{} of $10$-best lists generated using these methods.
Figure~\ref{fig:usual-beam} also depicts the \certainty{} and \variety{} of $n$-best lists generated using different beam sizes.

\begin{table}[t]
    \small
    \tabcolsep 5pt
    \centering
    \begin{tabular}{lccccc}
        \toprule
                      & \multicolumn{2}{c}{\certainty} & & \multicolumn{2}{c}{\variety{}} \\
        \cmidrule{2-3}\cmidrule{5-6}
        Model         & \entq{} & \cntq{} & & \entq{} & \cntq{} \\ \midrule
        Blender 400M  & 0.806   & 0.747   & & 0.780   & 0.775 \\
        Blender 1B    & 0.832   & 0.752   & & 0.832   & 0.753 \\
        Blender 3B    & 0.856   & 0.768   & & 0.824   & 0.737 \\
        DialoGPT 345M & 0.938   & 0.917   & & 0.750   & 0.669 \\
        DialoGPT 762M & 0.883   & 0.918   & & 0.671   & 0.713 \\ \bottomrule
    \end{tabular}
    \caption{\certainty{} and \variety{} of $10$-best lists using beam search with beam size $B=10$.}
    \label{tab:usual-beam10}
\end{table}

\paragraph{\certainty{}.}
Table~\ref{tab:usual-beam10} illustrates that in approximately $10\%$ of \cntq{}-type inputs, even the highest scoring model generates $10$-best lists full of contradictory responses.
Even with a perfect response filter, the models are unable to provide noncontradictory answers to these questions.
It should be emphasized that the error rate is not low, given that the inputs are polar questions with highly restricted viable responses.
Expanding the beam size can increase the number of $n$-best lists with at least one noncontradictory response.
Indeed, increasing the beam size enhances the \certainty{} ((a) and (b) in Figure~\ref{fig:usual-beam}).
By increasing $B$ to $40$, the \certainty{} of using DialoGPT $345$M for both \entq{}- and \cntq{}-type inputs achieve $1.0$. 

\paragraph{\variety{}.}
With $B=10$, all the models' \variety{} are more than $0.5$ (chance rate) (Table~\ref{tab:usual-beam10}).
Therefore, rather than being fully random, the models generate $n$-best lists with a degree of directionality toward avoiding contradictions.
However, increasing the size of beam reduces the \variety{} ((c) and (d) in Figure~\ref{fig:usual-beam}), resulting in lower output quality.
For example, the \variety{} of DialoGPT $345$M with $B=40$ for \cntq{}-type inputs (a model with \certainty{} of $1.0$ for both \entq{}- and \cntq{}-type inputs) decreases to $0.58$.

\paragraph{Overall.}
In terms of avoiding contradiction, our analytical framework demonstrated the features of the $n$-best lists of the beam search.
The \certainty{} did not achieve $1.0$ in the commonly used configuration ($B=10$).
When the beam size is increased, the \certainty{} increases to $1.0$, whereas the \variety{} reduces dramatically.
These results show the trade-off between \certainty{} and \variety{} as a function of beam size; in this example, we found constraints in obtaining high \certainty{} and \variety{} with beam search.
Furthermore, it is found that the \certainty{} obtained using DialoGPT is greater than that obtained using Blender, whereas the opposite is true for \variety{}, suggesting that various models behave differently in terms of \certainty{} and \variety{}.
This study emphasizes the significance of examining the \certainty{} and \variety{} of each model.

\subsection{Analysis of \textit{n}-best by various techniques}
\label{subsec:analysis-variousmethod}

\paragraph{How to achieve high \certainty{} and \variety{}?}
One method to increase \certainty{} is to generate $n$-best lists with a wider range of responses, such that each $n$-best list is guaranteed to contain a specific number of noncontradictory responses.
The diverse beam search (DBS)~\cite{vijayakumar:aaai2018:diverse} and nucleus sampling (NS)~\cite{holtzman:iclr2020:nucleus} methods are used to construct such $n$-best lists.
Furthermore, \citet{li:acl2020:dontsaythat} recently proposed models that use unlikelihood (UL) training to assign low probabilities to contradict responses.
Using these models to generate $n$-best lists will almost certainly enhance both \certainty{} and \variety{}.
We assess the $n$-best lists generated using these three strategies to see how much these techniques enhance \certainty{} and \variety{} ($n$-best lists generated using DBS and NS, and $n$-best lists generated using beam search together with the UL training).
Appendix~\ref{sec:details-analysis} contains a description of the techniques used for this analysis.

\begin{table}[t]
    \small
    \centering
    \begin{tabular}{lccccc}
        \toprule
                         & \multicolumn{2}{c}{\certainty} & & \multicolumn{2}{c}{\variety{}} \\
        \cmidrule{2-3}\cmidrule{5-6}
        Technique            & \entq{} & \cntq{} & & \entq{} & \cntq{} \\ \midrule
        BS               & 0.856   & 0.768   & & 0.824   & 0.737   \\
        DBS              & 0.999   & 0.981   & & 0.758   & 0.478   \\
        NS               & 1.000   & 0.994   & & 0.755   & 0.462   \\
        UL ($\alpha =0$)  & 1.000   & 0.996   & & 0.406   & 0.759   \\
        UL ($\alpha =1$)  & 0.943   & 0.900   & & 0.920   & 0.938   \\
        UL ($\alpha =10$) & 0.910   & 0.937   & & 0.969   & 0.968   \\\bottomrule
    \end{tabular}
    \caption{\certainty{} and \variety{} of $10$-best lists using various techniques with Blender $3$B.}
    \label{tab:various-beam10}
\end{table}

\paragraph{Result.}
Table~\ref{tab:various-beam10} displays the \certainty{} and \variety{} of the $10$-best lists generated using BS, DBS, NS, and UL.%
    \footnote{For the BS, DBS, and UL, we obtained the $10$-best lists setting beam size to $10$. For the NS, we got the $10$-best lists by performing nucleus sampling ten times.}
The values of $\alpha$ show the degree of UL loss during fine-tuning.
Here UL with $\alpha=0$ used the response generation model fine-tuned with maximum likelihood in the same training settings as those used for UL with $\alpha>0$.
Thus, note that comparing UL with $\alpha=0$ and $\alpha>0$ allows a fair comparison between likelihood and unlikelihood training.
The results reveal the properties of the $n$-best lists obtained for the three techniques, as well as the extent to which the techniques increase \certainty{} and \variety{}.
The \certainty{} obtained using the DBS and NS method reach $1.0$ for significantly lower search sizes than that for the BS to attain a \certainty{} of $1.0$; the \variety{} for \cntq{}-type inputs are less than $0.5$ (chance rate).
Thus, using the DBS and NS methods efficiently improves \certainty{} compared with the results obtained using the beam search; nevertheless, the methods do not simultaneously attain high \certainty{} and \variety{}.
However, the \certainty{} obtained using UL with $\alpha>0$ are greater than those obtained using the BS, and this was accomplished while maintaining higher \variety{} than those obtained using the BS and UL with $\alpha=0$ (likelihood training).
Our findings show that generation models are advancing toward high \certainty{} and \variety{}, which is particularly true for the recently proposed UL loss method.
Despite the highly restricted viable responses, i.e., \textit{yes} or \textit{no}, the \certainty{} obtained using UL with $\alpha>0$ does not reach $1.0$.
Thus, we conclude that there is still room for improvement in $n$-best list generation in terms of avoiding contradiction.


\section{Conclusion}
\label{sec:conclusion}
Based on the recent development of contradiction detectors, removing contradictory candidates from models' $n$-best lists is a practical method for avoiding contradiction.
In this method, the consistency of all candidates in the $n$-best lists substantially affects whether the final outputs are contradictory.

We quantitatively examined the properties of the $n$-best lists in terms of avoiding contradiction, using polar-typed questions as analytical inputs.
We demonstrated that the proposed framework exhibits the properties of $n$-best lists based on \certainty{} and \variety{}.
\certainty{} determines whether an n-best list has at least one noncontradictory response, whereas \variety{} evaluates how many noncontradictory responses each n-best list has.
The results, particularly, demonstrated the present limitations on achieving high \certainty{} and \variety{} when using the well-established beam search method.
In addition, our method emphasizes the improvements in \certainty{} and \variety{} achieved by recently proposed response generation strategies.

Our approach, which analyzes models' $n$-best lists based on \certainty{} and \variety{}, can be applied to any response generation problem, not just polar-typed response generation, which will be future work.


\section*{Acknowledgments}
We would like to thank all anonymous reviewers for their insightful comments.
We also thank Ana Brassard and Yosuke Kishinami for their valuable feedback and support.
This work was partly supported by JSPS KAKENHI Grant Numbers JP21J22383, JP22K17943, JST Moonshot R\&D Grant Number JPMJMS2011, and a Bilateral Joint Research Program between RIKEN AIP Center and Tohoku University.


\bibliography{references}

\begin{thebibliography}{21}
\expandafter\ifx\csname natexlab\endcsname\relax\def\natexlab#1{#1}\fi

\bibitem[{Adiwardana et~al.(2020)Adiwardana, Luong, So, Hall, Fiedel,
  Thoppilan, Yang, Kulshreshtha, Nemade, Lu, and
  Le}]{adiwardana:arxiv2020:meena}
Daniel Adiwardana, Minh-Thang Luong, David~R. So, Jamie Hall, Noah Fiedel,
  Romal Thoppilan, Zi~Yang, Apoorv Kulshreshtha, Gaurav Nemade, Yifeng Lu, and
  Quoc~V. Le. 2020.
\newblock \href {https://arxiv.org/abs/2001.09977v3} {Towards a human-like
  open-domain chatbot}.
\newblock In \emph{{arXiv} preprint {arXiv}:2001.09977}.

\bibitem[{Higashinaka et~al.(2015)Higashinaka, Funakoshi, Araki, Tsukahara,
  Kobayashi, and Mizukami}]{higashinaka:sigdial2015:taxonomy}
Ryuichiro Higashinaka, Kotaro Funakoshi, Masahiro Araki, Hiroshi Tsukahara,
  Yuka Kobayashi, and Masahiro Mizukami. 2015.
\newblock \href {https://doi.org/10.18653/v1/w15-4611} {Towards taxonomy of
  errors in chat-oriented dialogue systems}.
\newblock In \emph{Proceedings of the 16th annual meeting of the special
  interest group on discourse and dialogue ({SIGDIAL})}, pages 87--95.

\bibitem[{Holtzman et~al.(2020)Holtzman, Buys, Du, Forbes, and
  Choi}]{holtzman:iclr2020:nucleus}
Ari Holtzman, Jan Buys, Li~Du, Maxwell Forbes, and Yejin Choi. 2020.
\newblock \href {https://iclr.cc/virtual_2020/poster_rygGQyrFvH.html} {The
  {Curious} {Case} of {Neural} {Text} {Degeneration}}.
\newblock In \emph{Proceedings of the eighth international conference on
  learning representations ({ICLR})}.

\bibitem[{Honnibal and Montani(2017)}]{spacy2}
Matthew Honnibal and Ines Montani. 2017.
\newblock {spaCy} 2: {Natural} language understanding with {Bloom} embeddings,
  convolutional neural networks and incremental parsing.

\bibitem[{Kim et~al.(2020)Kim, Kim, and
  Kim}]{kim:emnlp2020:will-i-sound-like-me}
Hyunwoo Kim, Byeongchang Kim, and Gunhee Kim. 2020.
\newblock \href {https://doi.org/10.18653/v1/2020.emnlp-main.65} {Will {I}
  sound like me? {Improving} persona consistency in dialogues through pragmatic
  self-consciousness}.
\newblock In \emph{Proceedings of the 2020 conference on empirical methods in
  natural language processing ({EMNLP})}, pages 904--916.

\bibitem[{Li et~al.(2020)Li, Roller, Kulikov, Welleck, Boureau, Cho, and
  Weston}]{li:acl2020:dontsaythat}
Margaret Li, Stephen Roller, Ilia Kulikov, Sean Welleck, Y-Lan Boureau,
  Kyunghyun Cho, and Jason Weston. 2020.
\newblock \href {https://doi.org/10.18653/v1/2020.acl-main.428} {Don’t say
  that! {Making} inconsistent dialogue unlikely with unlikelihood training}.
\newblock In \emph{Proceedings of the 58th annual meeting of the association
  for computational linguistics ({ACL})}, pages 4715--4728.

\bibitem[{Li et~al.(2021)Li, Zhang, Fei, Feng, and
  Zhou}]{li:acl2021:addressing}
Zekang Li, Jinchao Zhang, Zhengcong Fei, Yang Feng, and Jie Zhou. 2021.
\newblock \href {https://doi.org/10.18653/v1/2021.findings-acl.91} {Addressing
  {Inquiries} about {History}: {An} {Efficient} and {Practical} {Framework} for
  {Evaluating} {Open}-domain {Chatbot} {Consistency}}.
\newblock In \emph{Findings of the joint conference of the 59th annual meeting
  of the association for computational linguistics and the 11th international
  joint conference on natural language processing ({ACL}-{IJCNLP})}, pages
  1057--1067.

\bibitem[{Liu et~al.(2019)Liu, Ott, Goyal, Du, Joshi, Chen, Levy, Lewis,
  Zettlemoyer, and Stoyanov}]{liu:arxiv2019:roberta}
Yinhan Liu, Myle Ott, Naman Goyal, Jingfei Du, Mandar Joshi, Danqi Chen, Omer
  Levy, Mike Lewis, Luke Zettlemoyer, and Veselin Stoyanov. 2019.
\newblock \href {http://arxiv.org/abs/1907.11692} {{RoBERTa}: {A} {Robustly}
  {Optimized} {BERT} {Pretraining} {Approach}}.
\newblock In \emph{{arXiv} preprint {arXiv}:1907.11692}.

\bibitem[{Louis et~al.(2020)Louis, Roth, and Radlinski}]{louis:emnlp2020:circa}
Annie Louis, Dan Roth, and Filip Radlinski. 2020.
\newblock \href {https://doi.org/10.18653/v1/2020.emnlp-main.601} {“{I}'d
  rather just go to bed”: {Understanding} {Indirect} {Answers}}.
\newblock In \emph{Proceedings of the 2020 conference on empirical methods in
  natural language processing ({EMNLP})}, pages 7411--7425.

\bibitem[{Miller et~al.(2017)Miller, Feng, Fisch, Lu, Batra, Bordes, Parikh,
  and Weston}]{miller:emnlp2017demo:ParlAI}
Alexander~H. Miller, Will Feng, Adam Fisch, Jiasen Lu, Dhruv Batra, Antoine
  Bordes, Devi Parikh, and Jason Weston. 2017.
\newblock \href {https://doi.org/10.18653/v1/D17-2014} {{ParlAI}: {A} dialog
  research software platform}.
\newblock In \emph{Proceedings of the 2017 conference on empirical methods in
  natural language processing ({EMNLP}): {System} demonstrations}, pages
  79--84.

\bibitem[{Nie et~al.(2020)Nie, Williamson, Bansal, Kiela, and
  Weston}]{nie:acl2020:i-like-fish}
Yixin Nie, Mary Williamson, Mohit Bansal, Douwe Kiela, and Jason Weston. 2020.
\newblock \href {https://doi.org/10.18653/v1/2021.acl-long.134} {I like fish,
  especially dolphins: {Addressing} {Contradictions} in {Dialogue} {Modeling}}.
\newblock In \emph{Proceedings of the 59th annual meeting of the association
  for computational linguistics ({ACL})}, pages 1699--1713.

\bibitem[{Noh(1998)}]{noh:lp1998:echo}
Eun-Ju Noh. 1998.
\newblock \href {https://doi.org/10.1023/A:1005361528891} {Echo {Questions}:
  {Metarepresentation} and {Pragmatic} {Enrichment}}.
\newblock \emph{Linguistics and Philosophy}, 21(6):603--628.

\bibitem[{Roller et~al.(2021)Roller, Dinan, Goyal, Ju, Williamson, Liu, Xu,
  Ott, Shuster, Smith, Boureau, and Weston}]{roller:eacl2021:blenderbot}
Stephen Roller, Emily Dinan, Naman Goyal, Da~Ju, Mary Williamson, Yinhan Liu,
  Jing Xu, Myle Ott, Kurt Shuster, Eric~M. Smith, Y-Lan Boureau, and Jason
  Weston. 2021.
\newblock \href {https://www.aclweb.org/anthology/2021.eacl-main.24/} {Recipes
  for building an open-domain chatbot}.
\newblock In \emph{Proceedings of the 16th conference of the european chapter
  of the association for computational linguistics: {Main} volume ({EACL})},
  pages 300--325.

\bibitem[{Serban et~al.(2016)Serban, Sordoni, Bengio, Courville, and
  Pineau}]{serban:aaai2016:HRED}
Iulian~Vlad Serban, Alessandro Sordoni, Yoshua Bengio, Aaron Courville, and
  Joelle Pineau. 2016.
\newblock \href {https://arxiv.org/abs/1507.04808v3} {Building end-to-end
  dialogue systems using generative hierarchical neural network models}.
\newblock In \emph{Proceedings of the 30th {AAAI} conference on artificial
  intelligence ({AAAI}-16)}, pages 3776--3783.

\bibitem[{Sordoni et~al.(2015)Sordoni, Galley, Auli, Brockett, Ji, Mitchell,
  Nie, Gao, and Dolan}]{sordoni:naacl2015:gen-context-sensitive}
Alessandro Sordoni, Michel Galley, Michael Auli, Chris Brockett, Yangfeng Ji,
  Margaret Mitchell, Jian-Yun Nie, Jianfeng Gao, and Bill Dolan. 2015.
\newblock \href {https://doi.org/10.3115/v1/n15-1020} {A neural network
  approach to context-sensitive generation of conversational responses}.
\newblock In \emph{Proceedings of the 2015 conference of the north american
  chapter of the association for computational linguistics: {Human} language
  technologies ({NAACL}-{HLT})}, pages 196--205.

\bibitem[{Vijayakumar et~al.(2016)Vijayakumar, Cogswell, Selvaraju, Sun, Lee,
  Crandall, and Batra}]{vijayakumar:aaai2018:diverse}
Ashwin~K. Vijayakumar, Michael Cogswell, Ramprasaath~R. Selvaraju, Qing Sun,
  Stefan Lee, David Crandall, and Dhruv Batra. 2016.
\newblock \href {https://openreview.net/forum?id=HJV1zP5xg} {Diverse {Beam}
  {Search} for {Improved} {Description} of {Complex} {Scenes}}.
\newblock In \emph{Proceedings of the 32nd {AAAI} conference on artificial
  intelligence ({AAAI}-18)}.

\bibitem[{Vinyals and Le(2015)}]{vinyals:icml2015ws:neuralconv}
Oriol Vinyals and Quoc~V. Le. 2015.
\newblock \href {http://arxiv.org/abs/1506.05869v3} {A neural conversational
  model}.
\newblock In \emph{Proceedings of the 31st international conference on machine
  learning ({ICML}) deep learning workshop}.

\bibitem[{Welleck et~al.(2020)Welleck, Kulikov, Roller, Dinan, Cho, and
  Weston}]{welleck:iclr2020:unlikelihood}
Sean Welleck, Ilia Kulikov, Stephen Roller, Emily Dinan, Kyunghyun Cho, and
  Jason Weston. 2020.
\newblock \href {https://iclr.cc/virtual_2020/poster_SJeYe0NtvH.html} {Neural
  {Text} {Generation} {With} {Unlikelihood} {Training}}.
\newblock In \emph{Proceedings of the eighth international conference on
  learning representations ({ICLR})}.

\bibitem[{Williams et~al.(2018)Williams, Nangia, and
  Bowman}]{williams:naacl2018:challengecorpus}
Adina Williams, Nikita Nangia, and Samuel~R. Bowman. 2018.
\newblock \href {https://doi.org/10.18653/v1/n18-1101} {A broad-coverage
  challenge corpus for sentence understanding through inference}.
\newblock In \emph{Proceedings of the 2018 conference of the north american
  chapter of the association for computational linguistics: {Human} language
  technologies ({NAACL}-{HLT})}, volume~1, pages 1112--1122.

\bibitem[{Wolf et~al.(2020)Wolf, Debut, Sanh, Chaumond, Delangue, Moi, Cistac,
  Rault, Louf, Funtowicz, Davison, Shleifer, von Platen, Ma, Jernite, Plu, Xu,
  Scao, Gugger, Drame, Lhoest, and Rush}]{wolf:emnlp2020:transformers}
Thomas Wolf, Lysandre Debut, Victor Sanh, Julien Chaumond, Clement Delangue,
  Anthony Moi, Pierric Cistac, Tim Rault, Rémi Louf, Morgan Funtowicz, Joe
  Davison, Sam Shleifer, Patrick von Platen, Clara Ma, Yacine Jernite, Julien
  Plu, Canwen Xu, Teven~Le Scao, Sylvain Gugger, Mariama Drame, Quentin Lhoest,
  and Alexander~M. Rush. 2020.
\newblock \href {https://www.aclweb.org/anthology/2020.emnlp-demos.6}
  {Transformers: {State}-of-the-art natural language processing}.
\newblock In \emph{Proceedings of the 2020 conference on empirical methods in
  natural language processing ({EMNLP}): {System} demonstrations}, pages
  38--45.

\bibitem[{Zhang et~al.(2020)Zhang, Sun, Galley, Chen, Brockett, Gao, Gao, Liu,
  and Dolan}]{zhang:acl2020demo:dialogpt}
Yizhe Zhang, Siqi Sun, Michel Galley, Yen-Chun Chen, Chris Brockett, Xiang Gao,
  Jianfeng Gao, Jingjing Liu, and Bill Dolan. 2020.
\newblock \href {https://doi.org/10.18653/v1/2020.acl-demos.30} {{DIALOGPT} :
  {Large}-scale generative pre-training for conversational response
  generation}.
\newblock In \emph{Proceedings of the 58th annual meeting of the association
  for computational linguistics ({ACL}): {System} demonstrations}, pages
  270--278.

\end{thebibliography}
\bibliographystyle{acl_natbib}

\clearpage

\appendix

\section{Details of transforming NLI data}
\label{sec:writing-rule}
As described in Section~\ref{subsec:input-preparation}, we obtain an analytical input from the NLI dataset.
Specifically, we convert the hypothesis sentence of an NLI sample into a yes-no question.
We describe the procedure as follows:
\begin{enumerate}
    \item Detect the first verb of a sentence.
    \item Move the verb to the beginning of the sentence, or put one of \{\textit{Do}, \textit{Does}, \textit{Did}\} at the front of the sentence, changing the verb back to its base (e.g., \textit{made} $\rightarrow$ \textit{make}).
    \item Change first-person pronouns to second-person pronouns and second-person pronouns to first-person pronouns (e.g., \textit{my} $\rightarrow$ \textit{your}).
    \item Change the punctuation mark at the end of the sentence to a question mark.
\end{enumerate}
We used spaCy (\texttt{en\_core\_web\_sm})~\cite{spacy2} to detect the verbs of hypothesis sentences.
We did not use NLI samples with syntactically complex hypothesis sentences, such as those containing coordinating conjunctions, to avoid obtaining ungrammatical inputs.
Further details are provided in our source codes.%
    \footnote{\url{https://github.com/shiki-sato/nbest-contradiction-analysis}}

\section{Details of yes-no classifier}
\label{sec:classifier}

\paragraph{Training settings.}
On the Circa dataset, we fine-tuned the pretrained RoBERTa (\texttt{roberta-large}) implemented by Hugging Face~\cite{wolf:emnlp2020:transformers}.
We divided the dataset at random into train$:$valid $= 8:2$.
The other training parameters were identical to those used by \citet{louis:emnlp2020:circa}.

\paragraph{Performance of classifier.}
To investigate the performance of the classifier, we measured the classification accuracy.
First, we manually labeled the top-$1$ responses in the $10$-best lists generated by the analysis presented in Section~\ref{subsec:analysis-beamsearch} with one of the two following labels: \textit{Contradictory} or \textit{Noncontradictory}.
The accuracy with which the automated evaluation categorized the labeled responses was then evaluated.
We selected $500$ responses%
\footnote{$100$ responses generated by each of $5$ generation models.}
from $50$ \entq{}/\cntq{} inputs drawn at random from our test for the evaluation. 
The classifier classified $433$/$500$ responses (see Appendix~\ref{sec:details-analysis}), and the accuracy was $0.921$.
Some examples of the classification are shown in Table~\ref{tab:classification-examples}.
The classifier correctly detected the contradiction in the model response using an indirect expression, in Example~$1$.
However, in Example~$2$, the classifier failed to detect the contradiction of the model response, having both a noncontradictory direct expression (``No'') and a contradictory indirect expression (the part of the response after ``No'').
We found that the classifier tended to misclassify model responses containing the contradictions with themselves, such as Example~$2$.

\begin{table}[t]
    \small
    \tabcolsep 2.5pt
    \begin{tabular}{rp{50mm}} 
        \toprule
            History:         & and we didn't ever call it uh Cokes and such you know we call it soda. \\
            Message:         & Don't you always call it Coke? \\
            Model Response:  & We call it coke. \\
        \midrule
            Human Label:     & \textbf{Contradictory} \\
            Predicted Label: & \textbf{Contradictory} \\
        \bottomrule
    \end{tabular}
    \vspace{-2mm}
    \caption*{(a) Example $1$}
    \vspace{2mm}

    \begin{tabular}{rp{50mm}} 
        \toprule
            History:         & The buying a house was the last thing that i wanted to do. \\
            Message:         & Weren't you desperate to buy a house? \\
            Model Response:  & No, I just wanted to buy a house. \\
        \midrule
            Human Label:     & \textbf{Contradictory} \\
            Predicted Label: & \textbf{Non-contradictory} \\
        \bottomrule
    \end{tabular}
    \vspace{-2mm}
    \caption*{(a) Example $2$}
    \vspace{-1mm}
   \caption{Examples of the response classification results by the yes-no classifier. The model responses were generated by Blender $400$M using beam search with beam size $B=10$.}
    \label{tab:classification-examples}
\end{table}

\begin{table}[t]
    \small
    \centering
    \begin{tabular}{lcc}
        \toprule
        Model         & \entq{}   & \cntq{}             \\ \midrule
        Blender 400M  & 1331 / 2000 & \phantom{}1270 / 2000 \\
        Blender 1B    & 1413 / 2000 & \phantom{}1316 / 2000 \\
        Blender 3B    & 1566 / 2000 & \phantom{}1403 / 2000 \\
        DialoGPT 345M & 1126 / 2000 & \phantom{0}924 / 2000 \\
        DialoGPT 762M & 1044 / 2000 & \phantom{0}956 / 2000 \\ \bottomrule
    \end{tabular}
    \caption{Number of stimulus inputs analyzed to calculate the \certainty{} and \variety{} described in Table~\ref{tab:usual-beam10}.}
    \label{tab:num-eval-beam}
\end{table}

\begin{table}[t!]
    \small
    \centering
    \begin{tabular}{p{21.3mm}cc}
        \toprule
        Model             & \entq{}             & \cntq{}             \\ \midrule
        BS                & \phantom{}1566 / 2000 & \phantom{}1403 / 2000 \\
        DBS               & \phantom{0}991 / 2000 & \phantom{0}882 / 2000 \\
        NS                & \phantom{0}818 / 2000 & \phantom{0}684 / 2000 \\
        UL ($\alpha =0$)  & \phantom{}1914 / 2000 & \phantom{}1871 / 2000 \\
        UL ($\alpha =1$)  & \phantom{}1806 / 2000 & \phantom{}1887 / 2000 \\
        UL ($\alpha =10$) & \phantom{}1654 / 2000 & \phantom{}1811 / 2000 \\        
        \bottomrule
    \end{tabular}
    \caption{Number of stimulus inputs analyzed to calculate the \certainty{} and \variety{} described in Table~\ref{tab:various-beam10}.}
    \label{tab:num-eval-various}
\end{table}

\section{Details of experiments}
\label{sec:details-analysis}

\paragraph{Number of analyzed stimulus inputs.}
To simplify the analysis, we omitted from Section~\ref{sec:experiments} and Appendix~\ref{sec:classifier} the analytical inputs with one or more ambiguous responses in the $n$-best lists.
We defined ambiguous responses as those that were not identified by the classifier as either affirmations or refutations.%
    \footnote{Circa dataset has seven different labels such as ``Yes'' and ``Probably/sometimes yes.'' We regard the responses classified into ``In the middle'' or ``I am not sure'' as ambiguous ones.}
Table~\ref{tab:num-eval-beam} and Table~\ref{tab:num-eval-various} display the number of analytical inputs from the total of $2,000$ \entq{}/\cntq{} used for the two analyses in Section~\ref{sec:experiments}.

\paragraph{Generation model settings.}
\label{subsec:generation-settings}
In Section~\ref{sec:experiments} experiments, we used DialoGPT~\cite{zhang:acl2020demo:dialogpt} and Blender~\cite{roller:eacl2021:blenderbot} as response generation models.
We used the codes of ParlAI~\cite{miller:emnlp2017demo:ParlAI} with its default settings, except for
{\small \texttt{beam\_length\_penalty}$=0$} to generate responses.

\paragraph{Unlikelihood training settings.}
We used unlikelihood training with Blender $3$B for the study of Section~\ref{subsec:analysis-variousmethod}.
To use the unlikelihood training proposed by~\citet{li:acl2020:dontsaythat}, we require training data that includes the following three elements: input (here, history, and message), gold response, and negative response.
These training samples were created by altering the NLI data with entailing and contradicting hypotheses.%
\footnote{Note that we did not use the identical NLI samples to synthesize \entq{}/\cntq{}.}
Table~\ref{tab:ul-transform} displays the original NLI data and the transformed training samples.
One NLI data set yields four types of questions (PositiveQ$1$, PositiveQ$2$, NegativeQ$1$, and NegativeQ$2$).
We synthesized $8,\!000$ samples from $2,\!000$ NLI data and randomly divided them into $\mathrm{train}:\mathrm{valid} = 9:1$.
We tuned the learning rate $\{7.0\times10^{-4}, 7.0\times10^{-5}, 7.0\times10^{-6}, 7.0\times10^{-7}, 7.0\times10^{-8}\}$ and the number of warmup updates $\{50, 100\}$ for each $\alpha=\{0,1,10\}$ for training.
The rest of the training parameters are identical to those used by~\citet{roller:eacl2021:blenderbot}.
It is worth noting that we only trained the models marked as UL in Section~\ref{subsec:analysis-variousmethod} on these transformed data.

\newpage
\begin{table}[H]
    \centering
    \small
    \tabcolsep 3pt
    \makeatletter
    \setlength{\@fptop}{10pt}
    \makeatother
    \begin{tabular}{rl} 
        \toprule
            Premise: & yeah i'm in North Carolina \\
            Hypothesis -- \textbf{Entailment}: & I'm in North Carolina. \\
            Hypothesis -- \textbf{Contradict}: & I'm in South Carolina. \\
        \bottomrule
    \end{tabular}
    \vspace{-2mm}
    \caption*{(a) Original NLI data}
    
    \begin{tabular}{rp{52mm}} 
        \\
            \multicolumn{2}{l}{\textbf{PositiveQ1}} \\
        \toprule
            \quad History: & Yeah I'm in North Carolina. \\
            \quad Message: & Are you in North Carolina? \\
            \quad Gold: & Yes, I'm in North Carolina. \\
            \quad Negative: & No, I'm in South Carolina. \\
        \bottomrule
        \\
            \multicolumn{2}{l}{\textbf{PositiveQ2}} \\ 
        \toprule
            \quad History: & Yeah I'm in North Carolina. \\
            \quad Message: & Are you in South Carolina? \\
            \quad Gold: & No, I'm in North Carolina. \\
            \quad Negative: & Yes, I'm in South Carolina. \\
        \bottomrule
        \\
            \multicolumn{2}{l}{\textbf{NegativeQ1}} \\ 
        \toprule
            \quad History: & Yeah I'm in North Carolina. \\
            \quad Message: & Aren't you in North Carolina? \\
            \quad Gold: & Yes, I'm in North Carolina. \\
            \quad Negative: & No, I'm in South Carolina. \\
        \bottomrule
        \\
            \multicolumn{2}{l}{\textbf{NegativeQ2}} \\ 
        \toprule
            \quad History: & Yeah I'm in North Carolina. \\
            \quad Message: & Aren't you in South Carolina? \\
            \quad Gold: & No, I'm in North Carolina. \\
            \quad Negative: & Yes, I'm in South Carolina. \\
        \bottomrule
    \end{tabular}
    \vspace{-2mm}
    \caption*{(b) Training samples for UL}
    \caption{Example of transforming (a) original NLI data to (b) training sample for UL. We synthesized four questions, i.e., PositiveQ$1$, PositiveQ$2$, NegativeQ$1$, and NegativeQ$2$, from each NLI sample.}
    \label{tab:ul-transform}
\end{table}

\end{document}